\pdfoutput=1

\documentclass[11pt]{article}

\usepackage{acl}
\usepackage{graphicx}
\usepackage{tikz}
\makeatletter
\newcommand{\xRightarrow}[2][]{\ext@arrow 0359\Rightarrowfill@{#1}{#2}}
\makeatother

\usepackage{times}
\usepackage{latexsym}

\usepackage{amsmath,amsfonts,bm}

\def\eqref#1{equation~\ref{#1}}

\def\1{\bm{1}}

\DeclareMathAlphabet{\mathsfit}{\encodingdefault}{\sfdefault}{m}{sl}
\SetMathAlphabet{\mathsfit}{bold}{\encodingdefault}{\sfdefault}{bx}{n}

\def\gE{{\mathcal{E}}}

\def\gG{{\mathcal{G}}}

\def\gR{{\mathcal{R}}}

\def\gV{{\mathcal{V}}}

\newcommand{\E}{\mathbb{E}}

\usepackage{amsmath}

\usepackage{booktabs}
\usepackage{algorithm}
\usepackage{algpseudocode}
\usepackage{hyperref}
\usepackage{url}

\usepackage{booktabs}
\usepackage{graphicx}
\usepackage{multirow}

\newcommand{\ans}[1]{[\![ #1 ]\!]}
\usepackage{proof}

\usepackage[T1]{fontenc}

\usepackage[utf8]{inputenc}

\usepackage{microtype}

\title{Advancing Abductive Reasoning in Knowledge Graphs through \\ Complex Logical Hypothesis Generation}

\author{Jiaxin Bai$^{1*}$,
Yicheng Wang$^{1*}$,
Tianshi Zheng$^1$,
Yue Guo$^1$,
Xin Liu$^2$,
and Yangqiu Song$^1$ \\
$^1$Hong Kong University of Science and Technology, Clear Water Bay, Hong Kong SAR\\
$^2$Amazon.com Inc, Palo Alto, USA\\
\{jbai, ywangmy, tzhengad, ygouar\}@connect.ust.hk, xliucr@amazon.com, yqsong@cse.ust.hk
}

\usepackage[normalem]{ulem}
\begin{document}
\maketitle

\def\thefootnote{*}\footnotetext{Equal Contribution}\def\thefootnote{\arabic{footnote}}


\begin{abstract}
Abductive reasoning is the process of making educated guesses to provide explanations for observations. 
Although many applications require the use of knowledge for explanations, the utilization of abductive reasoning in conjunction with structured knowledge, such as a knowledge graph, remains largely unexplored.
To fill this gap, this paper introduces the task of complex logical hypothesis generation, as an initial step towards abductive logical reasoning with KG.
In this task, we aim to generate a complex logical hypothesis so that it can explain a set of observations.
We find that the supervised trained generative model can generate logical hypotheses that are structurally closer to the reference hypothesis. 
However, when generalized to unseen observations, this training objective does not guarantee better hypothesis generation. 
To address this, we introduce the Reinforcement Learning from Knowledge Graph (RLF-KG) method, which minimizes differences between observations and conclusions drawn from generated hypotheses according to the KG.
Experiments show that, with RLF-KG's assistance, the generated hypotheses provide better explanations, and achieve state-of-the-art results on three widely used KGs.\footnote{https://github.com/HKUST-KnowComp/AbductiveKGR}
\end{abstract}

\section{Introduction}

Abductive reasoning plays a vital role in generating explanatory hypotheses for observed phenomena across various research domains \citep{abductive-learning}. It is a powerful tool with wide-ranging applications. For example, in cognitive neuroscience, reverse inference \citep{reverse-inference}, which is a form of abductive reasoning, is crucial for inferring the underlying cognitive processes based on observed brain activation patterns. Similarly, in clinical diagnostics, abductive reasoning is recognized as a key approach for studying cause-and-effect relationships \citep{clinical-diag}. Moreover, abductive reasoning is fundamental to the process of hypothesis generation in humans, animals, and computational machines \citep{cognition}. Its significance extends beyond these specific applications and encompasses diverse fields of study.
In this paper, we are focused on abductive reasoning with structured knowledge, specifically, a knowledge graph. 

A typical knowledge graph (KG) stores information about entities, like people, places, items, and their relations in graph structures.
Meanwhile, KG reasoning is the process that leverages knowledge graphs to infer or derive new information \citep{DBLP:journals/aiopen/ZhangCZKD21, DBLP:journals/corr/abs-2202-07412, DBLP:journals/tnn/JiPCMY22}. 
In recent years, various logical reasoning tasks are proposed over knowledge graph, for example, answering complex queries expressed in logical structure \citep{DBLP:conf/nips/HamiltonBZJL18, DBLP:conf/nips/RenL20}, or conducting logical rule mining \citep{DBLP:journals/vldb/GalarragaTHS15,DBLP:conf/semweb/HoSGKW18,DBLP:conf/ijcai/MeilickeCRS19}. 

\begin{figure*}[t]
\begin{center}
\includegraphics[width=\linewidth]{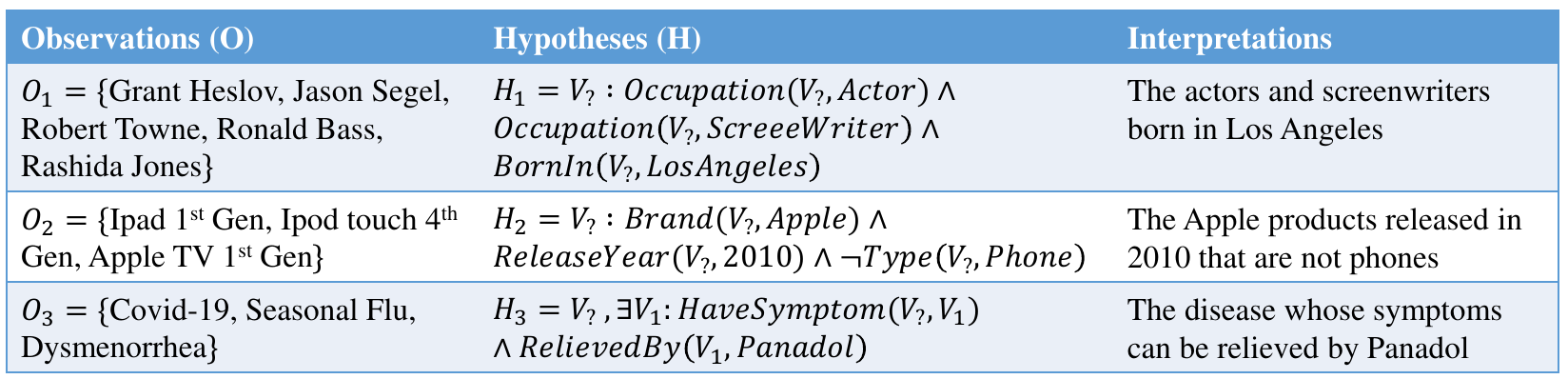}
\vspace{-1.0cm}
\end{center}
\caption{
This figure shows some examples of observations and inferred logical hypotheses, expressed with discrepancies in interpretations.}
\vspace{-0.4cm}
\label{fig:hypotheses_examples}
\end{figure*}

However, the abductive perspective of KG reasoning is crucial yet unexplored. 
Consider the first example in Figure \ref{fig:hypotheses_examples}, where observation \(O_1\) depicts five celebrities followed by a user on a social media platform.
The social network service provider is interested in using structured knowledge to explain the users' observed behavior. 
By leveraging a knowledge graph like Freebase \citep{freebase}, which contains basic information about these individuals, a complex logical hypothesis \(H_1\) can be derived suggesting that they are all actors and screenwriters born in Los Angeles.
In the second example from Figure \ref{fig:hypotheses_examples}, related to a user's interactions on an e-commerce platform ($O_2$), the structured hypothesis $H_2$ can explain the user's interest in $Apple$ products released in $2010$ excluding $phones$. 
The third example, dealing with medical diagnostics, presents three diseases (\(O_3\)). 
The corresponding hypothesis $H_3 $ indicates they are diseases $V_?$ with symptom $V_1$, and $V_1$ can be relieved by $Panadol$. 
From a general perspective, these problems illustrate how abductive reasoning with knowledge graphs seeks hypotheses that best explain given observations \citep{josephson1996abductive, thagard1997abductive}.

A straightforward approach to tackle this reasoning task is to employ a search-based method to explore potential hypotheses based on the given observation. However, this approach faces two significant challenges.
The first challenge arises from the incompleteness of knowledge graphs (KGs), as searching-based methods heavily rely on the availability of complete information. In practice, missing edges in KGs can negatively impact the performance of search-based methods \citep{DBLP:conf/nips/RenL20}.
The second challenge stems from the complexity of logically structured hypotheses. The search space for search-based methods becomes exponentially large when dealing with combinatorial numbers of candidate hypotheses. Consequently, the search-based method struggles to effectively and efficiently handle observations that require complex hypotheses for explanation.

To overcome these challenges, we propose a solution that leverages generative models within a supervised learning framework to generate logical hypotheses for given observations. Our approach involves sampling hypothesis-observation pairs from observed knowledge graphs \citep{DBLP:conf/iclr/RenHL20, bai2023sequential} and training a transformer-based generative model \citep{DBLP:conf/nips/VaswaniSPUJGKP17} using the teacher-forcing method.
However, a potential limitation of supervised training is that it primarily captures structural similarities, without necessarily guaranteeing improved explanations when applied to unseen observations. 
To address this, we introduce a technique called Reinforcement Learning from the Knowledge Graph (RLF-KG). It utilizes proximal policy optimization (PPO) \citep{ppo} to minimize the discrepancy between the observed evidence and the conclusion derived from the generated hypothesis. 
By incorporating reinforcement learning techniques, our approach aims to directly improve the explanatory capability of the generated hypotheses and ensure their effectiveness when generalized to unseen observations.

We evaluate the proposed methods for effectiveness and efficiency on three knowledge graphs: FB15k-237 \citep{DBLP:conf/acl-cvsc/ToutanovaC15}, WN18RR \citep{DBLP:conf/acl-cvsc/ToutanovaC15}, and DBpedia50 \citep{DBLP:conf/aaai/ShiW18}. The results consistently demonstrate the superiority of our approach over supervised generation baselines and search-based methods, as measured by two evaluation metrics across all three datasets. Our contributions can be summarized as follows:
\begin{itemize}
    \item We introduce the task of complex logical hypothesis generation, which aims to identify logical hypotheses that best explain a given set of observations. This task can be seen as a form of abductive reasoning with KGs.
    \item To address the challenges posed by the incompleteness of knowledge graphs and the complexity of logical hypotheses, we propose a generation-based method. This approach effectively handles these difficulties and enhances the quality of generated hypotheses.
    \item Additionally, we developed the Reinforcement Learning from the Knowledge Graph (RLF-KG) technique. By incorporating feedback from the knowledge graph, RLF-KG further improves the hypothesis generation model. It minimizes the discrepancies between the observations and the conclusions of the generated hypotheses, leading to more accurate and reliable results.
\end{itemize}

\section{Problem Formulation}
\label{sec:prob-form}

In this task, a knowledge graph is denoted as \(\gG = (\mathcal{V}, \mathcal{R})\), where \(\mathcal{V}\) is the set of vertices and \(\mathcal{R}\) is the set of relation types. Each \textit{relation type} \(r \in \mathcal{R}\) is a function \(: \mathcal{V} \times \mathcal{V} \to \{ \mathtt{true}, \mathtt{false} \}\), with $r(u, v) = \mathtt{true}$ indicating the existence of a directed $(u, r, v)$ from vertex \(u\) to vertex \(v\) of type \(r\) in the graph , and \(\mathtt{false}\) otherwise.

We adopt the open-world assumption of knowledge graphs \citep{drummond2006open}, treating missing edges as unknown rather than false. 
The reasoning model can only access the observed KG $\gG$, while the true KG $\mathcal{\bar{G}}$ is hidden and encompasses the observed graph $\gG$.

Abductive reasoning is a type of logical reasoning that involves making educated guesses to infer the most likely reasons for the observations \citep{josephson1996abductive, thagard1997abductive}. For further details on the distinctions between abductive, deductive, and inductive reasoning, refer to Appendix~\ref{app:clar-abd-ded-ind}.
In this work, we focus on a specific abductive reasoning type in the context of knowledge graphs. We first introduce some concepts in this context.

An \textit{observation} is a set $O$ of entities in $\gV$. 
A logical \textit{hypothesis} \(H\) on a graph \(\gG = (\mathcal{V}, \mathcal{R})\) is defined as a predicate of a variable vertex \(V_{?}\) in first-order logical form, including existential quantifiers, \texttt{AND} ($\land$), \texttt{OR} ($\lor$), and \texttt{NOT} ($\lnot$). The hypothesis can always be written in disjunctive normal form,
\begin{align}
H|_{\gG}(V_{?}) &= \exists V_1, \dots , V_k: e_1 \lor \dots \lor e_n,\\
e_i &= r_{i1} \land \dots \land r_{i m_i},
\end{align}
where each \(r_{ij}\) can take the forms:
$r_{ij} = r(v, V)$, $r_{ij} = \neg r(v, V)$, $r_{ij} = r(V, V')$, $r_{ij} = \neg r(V, V')$,  where \(v\) represents a fixed vertex, the \(V, V'\) represent variable vertices in \(\{ V_{?}, V_1, V_2, \dots, V_k \}\), and \(r\) is a relation type.

The subscript \(|_{\gG}\) denotes that the hypothesis is formulated based on the given graph \(\gG\). 
This means that all entities and relations in the hypothesis must exist in \(\gG\), and the domain for variable vertices is the entity set of  \(\gG\). For example, please refer to Appendix~~\ref{app:obse-hypo-eg}. The same hypothesis \(H\) can be applied to a different knowledge graph, \(\gG'\), provided that  \(\gG'\) includes the entities and edges present in \(H\).
When the context is clear or the hypothesis pertains to a general statement applicable to multiple knowledge graphs (e.g., observed and hidden graphs), the symbol \(H\) is used without the subscript.

\begin{figure*}
    \begin{minipage}{.4\linewidth}
        \includegraphics[width=\linewidth]{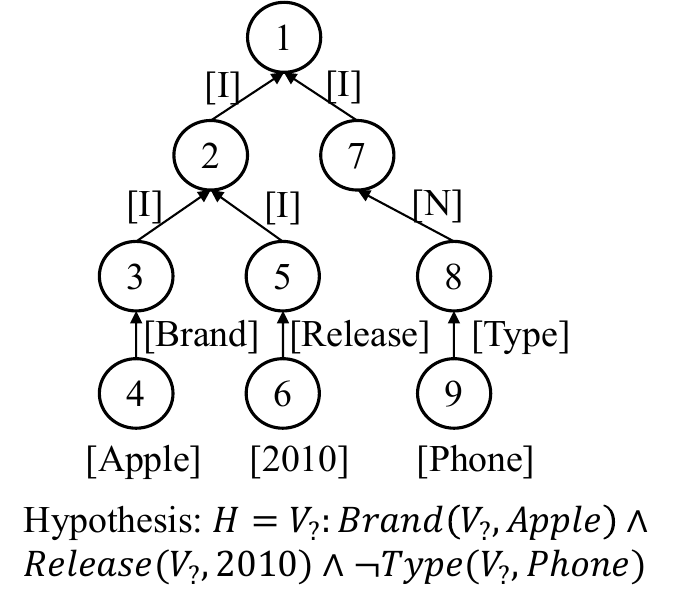}
    \end{minipage} 
    \(\iff\)
    \begin{minipage}{.4\textwidth}
        \begin{tabular}{c|c|c}
            \toprule
            \textbf{Nodes} & \textbf{Actions} & \textbf{Stack} \\
            \midrule
            1 & [I] & 1\\
            2 & [I] & 1,2\\
            3 & [Brand] & 1,2,3\\
            4 & [Apple] & 1,2,3,4 \(\to\) 1,2\\
            5 & [Release] & 1,2,5\\
            6 & [2010] & 1,2,5,6 \(\to\) 1\\
            7 & [N] & 1,7\\
            8 & [Type] & 1,7,8\\
            9 & [Phone] & 1,7,8,9 \(\to\) empty\\
            \bottomrule
        \end{tabular}
        Tokens: [I][I][Brand][Apple]

        [Release][2010][N][Type][Phone]
    \end{minipage}
    \vspace{-0.2cm}
    \caption{The figure demonstrates the tokenization process for hypotheses. We uniformly consider logical operations, relations, and entities as individual tokens, establishing a correspondence between the hypotheses and a sequence of tokens. For a more detailed explanation, please refer to the Appendix \ref{sec:algo-sampling}.}
    \label{fig:tokenization}
\end{figure*}

The \textit{conclusion} of the hypothesis \(H\) on a graph \(\gG\), denoted by \(\ans{H}_{\gG}\), is the set of entities for which \(H\) holds true on \(\gG\):
\begin{equation}
    \ans{H}_{\gG} = \{V_? \in \gG | H|_{\gG}(V_?) = \mathtt{true} \}.
\end{equation}

Suppose $O = \{v_1, v_2, ..., v_k\}$ represents an observation, \(\gG\) is the observed graph, and $\bar{\gG}$ is the hidden graph. Then \textit{abductive reasoning in knowledge graphs} aims to find the hypothesis \(H\) on \(\gG\)
whose conclusion on the hidden graph \(\bar{\gG}\), $\ans{H}_{\bar{\gG}}$, is  most similar to \(O\). Formally, the similarity is quantified using the Jaccard index, defined as:
\begin{equation}
    \label{eqn:ar-obj}
    \mathtt{Jaccard}( \ans{H}_{\bar{\gG}}, O) = \frac{|\ans{H}_{\bar{\gG}} \cap O|}{|\ans{H}_{\bar{\gG}} \cup O|}.
\end{equation}
In other words, the goal is to find a hypothesis \(H\) that maximizes \(\mathtt{Jaccard}( \ans{H}_{\bar{\gG}}, O)\).

\begin{figure*}[t]
\begin{center}
\includegraphics[width=0.9\linewidth]{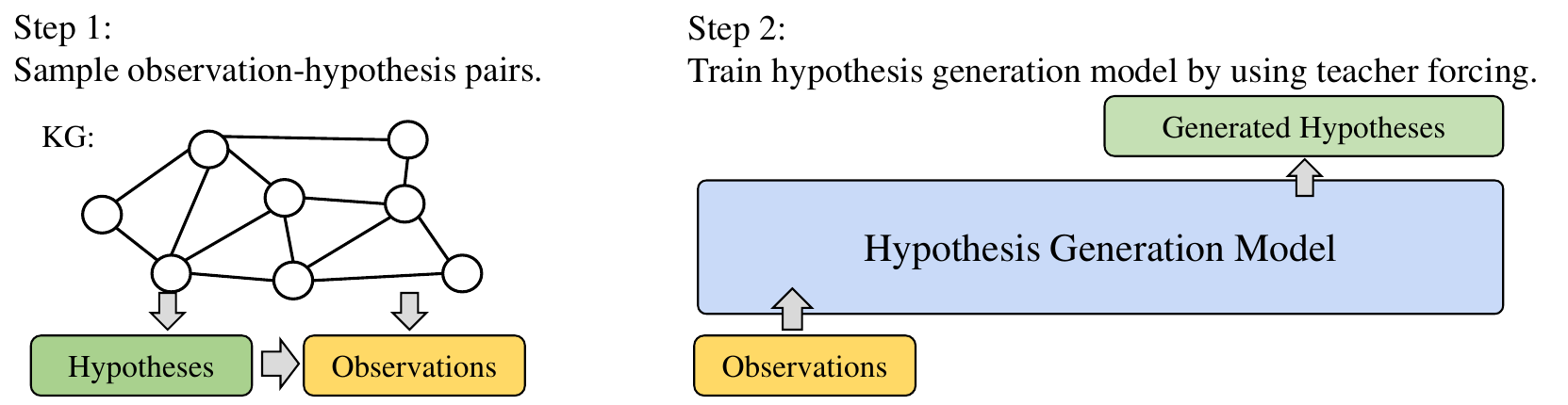}
\vspace{-0.4cm}
\end{center}
\caption{
The first two steps of training a hypothesis generation model:
In Step 1, we randomly sample logical hypotheses with diverse patterns and perform graph searches on the training graphs to obtain observations. These observations are then tokenized. 
In Step 2, a conditional generation model is trained to generate hypotheses based on given tokenized observations. 
}
\vspace{-0.3cm}
\label{fig:methods}
\end{figure*}

\section{Hypothesis Generation with RLF-KG}

\begin{figure*}[t]
\begin{center}
\includegraphics[width=\linewidth]{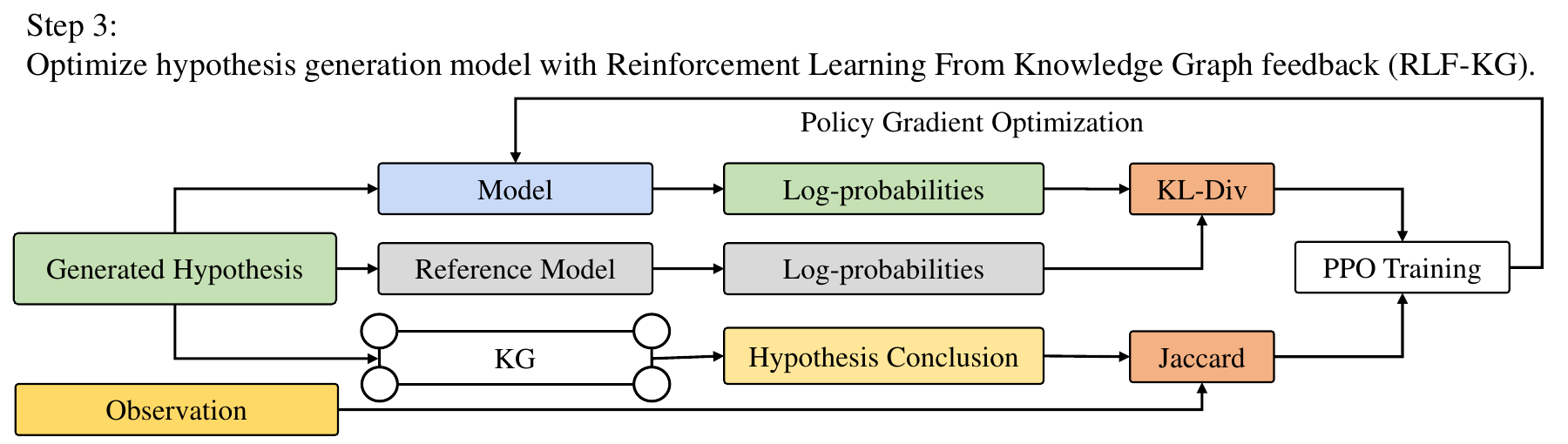}
\vspace{-0.8cm}
\end{center}
\caption{
In Step 3, we employ RLF-KG to encourage the model to generate hypotheses that align more closely with the given observations from the knowledge graph. RLF-KG helps improve the consistency between the generated hypotheses and the observed evidence in the knowledge graph.}
\vspace{-0.3cm}
\label{fig:rlfk}
\end{figure*}

Our methodology, referred to as reinforcement learning from the knowledge graph (RLF-KG), is depicted in Fig.~\ref{fig:rlfk} and involves the following steps:
(1) Randomly sample observation-hypothesis pairs from the knowledge graph. 
(2) Train a generative model to generate hypotheses from observations using the pairs. (3) Enhance the generative model using RLF-KG, leveraging reinforcement learning to minimize discrepancies between observations and generated hypotheses.

\subsection{Sampling Observation-Hypothesis Pairs}
\label{sec:sampling}

In the first step, we randomly sample hypotheses from the observed training knowledge graph. This process starts by randomly selecting a hypothesis, followed by conducting a graph search on the training graph to derive its conclusion, which is then treated as the observation corresponding to the hypothesis. Details of the hypothesis sampling algorithm are provided in Appendix~\ref{sec:algo-sampling}.

Then, we convert both hypotheses and observations into sequences suitable for the generative models. For observations, we standardize the order of the elements, ensuring that permutations of the same observation set yield identical outputs. Each entity in an observation is represented as a unique token, such as \texttt{[Apple]} and \texttt{[Phone]}, as shown in Figure \ref{fig:tokenization}, and is associated with an embedding. 

Since each hypothesis can be represented as a directed acyclic graph, for hypotheses, we use a representation inspired by action-based parsing, similar to approaches seen in other logical reasoning studies \citep{DBLP:conf/nips/HamiltonBZJL18, DBLP:conf/nips/RenL20, DBLP:conf/iclr/RenHL20}. This involves utilizing a depth-first search algorithm, described in Appendix \ref{sec:algo-sampling}, to generate a sequence of \textit{actions} that represents the content and structure of the graph. Logical operations such as intersection, union, and negation are denoted by special tokens \texttt{[I]}, \texttt{[U]}, and \texttt{[N]} respectively, following prior work \citep{bai2023sequential}. Relations and entities are similarly treated as unique tokens, for example, \texttt{[Brand]} and \texttt{[Apple]}. 

Furthermore, Algorithm \ref{alg:a2q} facilitates the reconstruction of a graph from an action sequence, serving as the detokenization process for hypotheses.

\subsection{Supervised Training of Hypothesis Generation Model}
\label{sec:supervised-training}

In the second step, we train a generative model using the sampled pairs. Let $\mathbf{o} = [o_1, o_2, ..., o_m ]$ represent the token sequences for observations, and $\mathbf{h} = [h_1, h_2, ..., h_n]$ for hypotheses. The loss for the generative model \(\rho\) on this example is based on the standard sequence modeling loss:
\begin{align}
     \mathcal{L} &= \log \rho(\mathbf{h} | \mathbf{o}) \\
     &= \log \sum_{i=1}^{n} \rho( h_i| \mathbf{o}, h_1,\dots, h_{i-1} ).
\end{align}
We utilize a standard transformer to implement the conditional generative model, employing two distinct approaches. The first approach follows the encoder-decoder architecture as described by \citet{DBLP:conf/nips/VaswaniSPUJGKP17}, where observation tokens are fed into the transformer encoder, and the shifted hypothesis tokens are input into the transformer decoder. The second approach involves concatenating the observation and hypothesis tokens and using a decoder-only transformer to generate hypotheses. Following the setup of these architectures, we train the model using supervised training techniques.

\subsection{Reinforcement Learning from Knowledge Graph Feedback (RLF-KG)}


During the supervised training process, the model learns to generate hypotheses that have similar structures to reference hypotheses. However, higher structural similarity towards reference answers does not necessarily guarantee the ability to generate logical explanations, especially when encountering unseen observations during training.

To address this limitation, in the third step, we employ reinforcement learning \citep{rlhf} in conjunction with knowledge graph feedback (RLF-KG) to enhance the trained conditional generation model \(\rho\). Let \(\gG_{\text{train}}\) represents the observed training graph, \(\mathbf{h}\) the hypothesis token sequence,  \(\mathbf{o}\) the observation token sequence, and \(H, O\) the corresponding hypothesis and observation, respectively. We select the reward to be the Jaccard similarity between \(O\) and and the conclusion \(\ans{H}_{\mathcal{G_\text{train}}}\), which is a reliable and information leakage-free approximation for the objective of the abductive reasoning task in Equation~\ref{eqn:ar-obj}. Formally, the reward function is defined as
\begin{equation}
   r(\mathbf{h}, \mathbf{o}) =  \mathtt{Jaccard}( \ans{H}_{\gG_{\text{train}}}, O) = \frac{| \ans{H}_{\gG_{\text{train}}} \cap O|}{|\ans{H}_{\gG_{\text{train}}} \cup O|}.
   \label{eq:reward}
\end{equation}

Following \citet{rlhf}, we treat the trained model \(\rho\) obtained from supervised training as the reference model, and initialize the model \(\pi\) to be optimized with  \(\rho\).
%
Then, we modify the reward function by incorporating a KL divergence penalty. This modification aims to discourage the model \(\pi\) from generating hypotheses that deviate excessively from the reference model. 

To train the model \(\pi\), we employ the proximal policy optimization (PPO) algorithm \citep{ppo}. The objective is to maximize the expected modified reward, given as follows:
\begin{equation}
\E_{\mathbf{o} \sim D, \mathbf{h} \sim \pi(\cdot | \mathbf{o})} \left[  r(\mathbf{h}, \mathbf{o}) - 
\beta \log \frac{\pi(\mathbf{h} | \mathbf{o})}{\rho(\mathbf{h} | \mathbf{o})}\right],
\end{equation}
where $ D $ the is training observation distribution and \(\pi(\cdot | \mathbf{o})\) is the conditional distribution of $\mathbf{h}$ on  $\mathbf{o}$ modeled by the model $\pi$. 
By utilizing PPO and the modified reward function, we can effectively guide the model \(\pi\) towards generating hypotheses that strike a balance between the similarity to the reference model and logical coherence. 


\section{Experiment}

\begin{figure*}[t]
\begin{center}
\includegraphics[width=\linewidth]{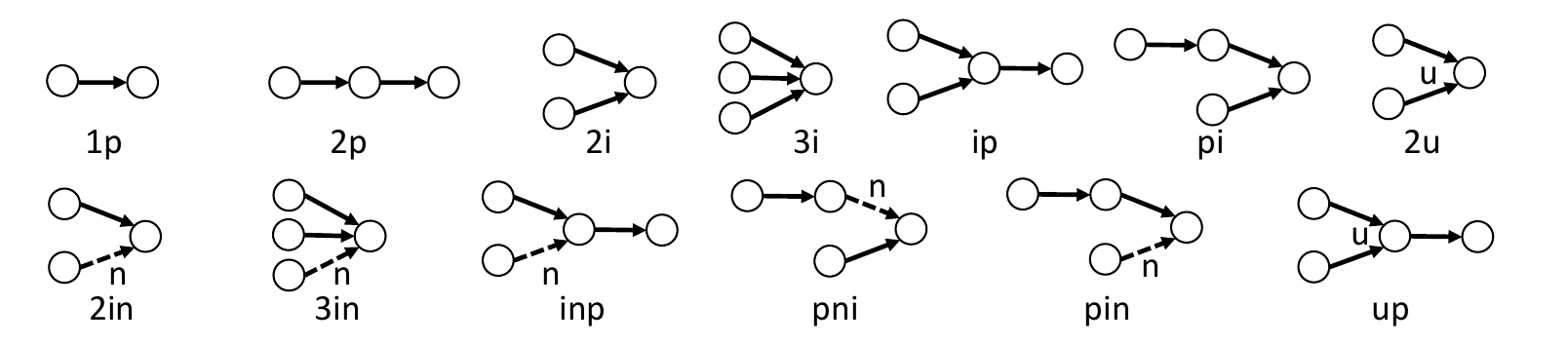}
\vspace{-1.1cm}
\end{center}
\caption{
Our task involves considering thirteen distinct types of logical hypotheses. Each hypothesis type corresponds to a specific type of query graph, which is utilized during the sampling process. By associating each hypothesis type with a corresponding query graph, we ensure that a diverse range of hypotheses is sampled.
}
\label{fig:hypotheses_types}
\vspace{-0.3cm}
\end{figure*}

\begin{table*}[t]
\begin{center}
\begin{tabular}{l|cc|ccc|c}
\toprule
\textbf{Dataset} & \textbf{Relations} & \textbf{Entities} & \textbf{Training} & \textbf{Validation} & \textbf{Testing} & \textbf{All Edges}\\
\midrule
FB15k-237 & 237 & 14,505 & 496,126 & 62,016 & 62,016 & 620,158\\
WN18RR & 11 & 40,559 & 148,132 & 18,516 & 18,516 & 185,164\\
DBpedia50 & 351 & 24,624 & 55,074 & 6,884 & 6,884 & 68,842\\

\bottomrule
\end{tabular}
\caption{This figure provides basic information about the three knowledge graphs utilized in our experiments. The graphs are divided into standard sets of training, validation, and testing edges to facilitate the evaluation process.}
\label{tab:KG_details}
\end{center}
\vspace{-0.5cm}
\end{table*}

We utilize three distinct knowledge graphs, namely FB15k-237 \citep{DBLP:conf/acl-cvsc/ToutanovaC15}, DBpedia50 \citep{DBLP:conf/aaai/ShiW18}, and WN18RR \citep{DBLP:conf/acl-cvsc/ToutanovaC15}, for our experiments. Table \ref{tab:KG_details} provides an overview of the number of training, evaluation, and testing edges, as well as the total number of nodes in each knowledge graph. 
To ensure consistency, we randomly partition the edges of these knowledge graphs into three sets - training, validation, and testing - using an 8:1:1 ratio. Consequently, we construct the training graph \(\gG_{\text{train}}\), validation graph \(\gG_{\text{valid}}\), and testing graph \(\gG_{\text{test}}\) by including the corresponding edges: training edges only, training + validation edges, and training + validation + testing edges, respectively.

Following the methodology outlined in Section~\ref{sec:supervised-training}, we proceed to sample pairs of observations and hypotheses. To ensure the quality and diversity of the samples, we impose certain constraints during the sampling process.
Firstly, we restrict the size of the observation sets to a maximum of thirty-two elements. This limitation is enforced ensuring that the observations remain manageable.
Additionally, specific constraints are applied to the validation and testing hypotheses. Each validation hypothesis must incorporate additional entities in the conclusion compared to the training graph, while each testing hypothesis must have additional entities in the conclusion compared to the validation graph. This progressive increase in entity complexity ensures a challenging evaluation setting. The statistics of queries sampled for datasets are detailed in Appendix~\ref{sec:queries-details}.

In line with previous work on KG reasoning \citep{DBLP:conf/nips/RenL20, DBLP:conf/iclr/RenHL20}, we utilize thirteen pre-defined logical patterns to sample the hypotheses. Eight of these patterns, known as existential positive first-order (EPFO) hypotheses (1p/2p/2u/3i/ip/up/2i/pi), do not involve negations. The remaining five patterns are negation hypotheses (2in/3in/inp/pni/pin), which incorporate negations. It is important to note that the generated hypotheses may or may not match the type of the reference hypothesis.
The structures of the hypotheses are visually presented in Figure \ref{fig:hypotheses_types}, and the corresponding numbers of samples drawn for each hypothesis type can be found in Table~\ref{tab:queries_details}.

\subsection{Evaluation Metric}

\paragraph{Jaccard Index} The quality of the generated hypothesis is primarily measured using the Jaccard index same as that defined for abductive reasoning in Section~\ref{sec:prob-form}, { but we treat the constructed test graph \(\gG_{\text{test}}\) as the hidden graph.}  It is important to note that the test graph contains ten percent of edges that were not observed during the training or validation stages. Formally, given an observation \(O\) and a generated hypothesis \(H\), we employ a graph search algorithm to determine the conclusion of \(H\) on \(\gG_{\text{test}}\), denoted as \(\ans{H}_{\gG_{\text{test}}}\). Then the Jaccard metric for evaluation is defined as 
\begin{equation}
    \mathtt{Jaccard}(\ans{H}_{\gG_{\text{test}}}, O) = 
    \frac{|  \ans{H}_{\gG_{\text{test}}}\cap  O|}{| \ans{H}_{\gG_{\text{test}}} \cup O |},
\end{equation}
quantifing the similarity between the conclusion \(\ans{H}_{\gG_{\text{test}}}\) and the observation \(O\).

\paragraph{Smatch Score}  Smatch \citep{DBLP:conf/acl/CaiK13} is originally designed for comparing semantic graphs but has been recognized as a suitable metric for evaluating complex logical queries, which can be treated as a specialized form of semantic graphs \citep{bai2023sequential}. In this task, a hypothesis can be represented as a graph, e.g. Fig.~\ref{fig:tokenization}, and we can transform it to be compatible with the semantic graph format.
The detailed computation of the Smatch score on hypothesis graphs is described in detail in Appendix \ref{sec:Smatch}. Intuitively, the Smatch score between the generated hypothesis \(H\) and the reference hypothesis \(H_{\text{ref}}\), denoted as \(S(H, H_{\text{ref}})\), quantifies the structural resemblance between the graphs corresponding to \(H\) and \(H_{\text{ref}}\) i.e., how much the nodes, edges, and the labels on them look the same between the two graphs.

\begin{table*}[t]
\begin{center}
\tiny
\begin{tabular}{@{}l|l|p{0.3cm}p{0.3cm}p{0.3cm}p{0.3cm}p{0.3cm}p{0.3cm}p{0.3cm}p{0.4cm}|p{0.3cm}p{0.3cm}p{0.3cm}p{0.3cm}p{0.4cm}|c@{}}
\toprule
\textbf{Dataset} & \textbf{Model} & \textbf{1p} & \textbf{2p} & \textbf{2i} & \textbf{3i} & \textbf{ip} & \textbf{pi} & \textbf{2u} & \textbf{up} & \textbf{2in} & \textbf{3in} & \textbf{pni} & \textbf{pin} & \textbf{inp} & \textbf{Ave.} \\ \midrule
\multirow{4}{*}{FB15k-237} 
& Encoder-Decoder & 0.626 & 0.617 & 0.551 & 0.513 & 0.576 & 0.493 & 0.818 & 0.613 & 0.532 & 0.451 & 0.499 & 0.529 & 0.533 & 0.565 \\
 & + RLF-KG & \textbf{0.855} & \textbf{0.711} & \textbf{0.661} & \textbf{0.595} & \textbf{0.715} & \textbf{0.608} & \textbf{0.776} & \textbf{0.698} & \textbf{0.670} & \textbf{0.530} & \textbf{0.617} & \textbf{0.590} & \textbf{0.637} & \textbf{0.666} \\ \cmidrule(l){2-16} 
 & Decoder-Only & 0.666 & 0.643 & 0.593 & 0.554 & 0.612 & 0.533 & 0.807 & 0.638 & 0.588 & 0.503 & 0.549 & 0.559 & 0.564 & 0.601 \\
 & + RLF-KG & \textbf{0.789} & \textbf{0.681} & \textbf{0.656} & \textbf{0.605} & \textbf{0.683} & \textbf{0.600} & \textbf{0.817} & \textbf{0.672} & \textbf{0.672} & \textbf{0.560} & \textbf{0.627} & \textbf{0.596} & \textbf{0.626} & \textbf{0.660} \\ \midrule
\multirow{4}{*}{WN18RR} 
& Encoder-Decoder & 0.793 & 0.734 & 0.692 & 0.692 & 0.797 & 0.627 & 0.763 & 0.690 & 0.707 & 0.694 & 0.704 & 0.653 & 0.664 & 0.708 \\
 & + RLF-KG & \textbf{0.850} & \textbf{0.778} & \textbf{0.765} & \textbf{0.763} & \textbf{0.854} & \textbf{0.685} & \textbf{0.767} & \textbf{0.719} & \textbf{0.743} & \textbf{0.732} & \textbf{0.738} & \textbf{0.682} & \textbf{0.710} & \textbf{0.753} \\ \cmidrule(l){2-16} 
 & Decoder-Only & 0.760 & 0.734 & 0.680 & 0.684 & 0.770 & 0.614 & 0.725 & 0.650 & 0.683 & 0.672 & 0.688 & 0.660 & 0.677 & 0.692 \\
 & + RLF-KG & \textbf{0.821} & \textbf{0.760} & \textbf{0.694} & \textbf{0.693} & \textbf{0.827} & \textbf{0.656} & \textbf{0.770} & \textbf{0.680} & \textbf{0.717} & \textbf{0.704} & \textbf{0.720} & \textbf{0.676} & \textbf{0.721} & \textbf{0.726} \\ \midrule
\multirow{4}{*}{DBpedia50} 
& Encoder-Decoder & 0.706 & 0.657 & 0.551 & 0.570 & 0.720 & 0.583 & 0.632 & 0.636 & 0.602 & 0.572 & 0.668 & 0.625 & 0.636 & 0.627 \\
 & + RLF-KG & \textbf{0.842} & \textbf{0.768} & \textbf{0.636} & \textbf{0.639} & \textbf{0.860} & \textbf{0.667} & \textbf{0.714} & \textbf{0.758} & \textbf{0.699} & \textbf{0.661} & \textbf{0.775} & \textbf{0.716} & \textbf{0.769} & \textbf{0.731} \\ \cmidrule(l){2-16} 
 & Decoder-Only & 0.739 & 0.692 & 0.426 & 0.434 & 0.771 & 0.527 & \textbf{0.654} & 0.688 & 0.602 & 0.563 & 0.663 & \textbf{0.640} & 0.701 & 0.623 \\
 & + RLF-KG & \textbf{0.777} & \textbf{0.701} & \textbf{0.470} & \textbf{0.475} & \textbf{0.821} & \textbf{0.534} & 0.646 & \textbf{0.702} & \textbf{0.626} & \textbf{0.575} & \textbf{0.696} & 0.626 & \textbf{0.713} & \textbf{0.643} \\ \bottomrule
\end{tabular}
\end{center}
\vspace{-0.4cm}
\caption{The detailed Jaccard performance of various methods.}
\label{tab:general_results}
\vspace{-0.2cm}
\end{table*}

\begin{figure*}[t]
\begin{center}
\includegraphics[width=\linewidth]{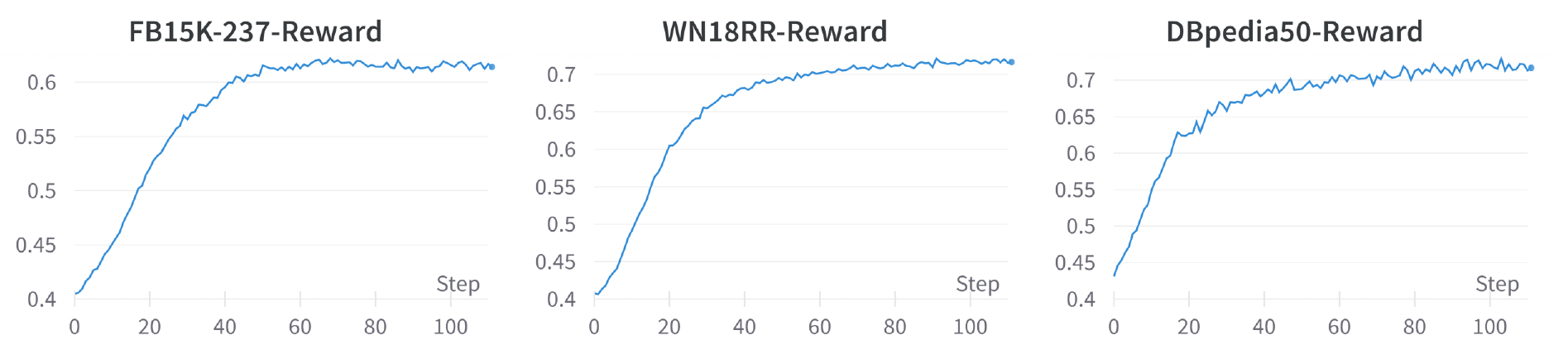}
\vspace{-1.1cm}
\end{center}
\caption{
The curve of the reward values of RLF-KG training over three different datasets. 
}
\label{fig:reward_figures}
\end{figure*}

\subsection{Experiment Details}

In this experiment, we use two transformer structures as the backbones of the generation model. 
For the encoder-decoder transformer structure, we use three encoder layers and three decoder layers. Each layer has eight attention heads with a hidden size of 512. Note that the positional encoding for the input observation sequence is disabled, as we believe that the order of the entities in the observation set does not matter. 
For the decoder-only structure, we use six layers, and the other hyperparameters are the same. 
In the supervised training process, we use AdamW optimizer and grid search to find hyper-parameters. 
For the encoder-decoder structure, the learning rate is $0.0001$ with the resulting batch size of  $768$, $640$, and $256$ for FB15k-237, WN18RR, and DBpedia, respectively. For the decoder-only structure, the learning rate is $0.00001$ with batch-size of $256$, $160$, and $160$ for FB15k-237, WN18RR, and DBpedia respectively, and linear warming up of $100$ steps. 
In the reinforcement learning process, we use the dynamic adjustment of the penalty coefficient $\beta$ \citep{DBLP:conf/nips/Ouyang0JAWMZASR22}. More detailed hyperparameters are shown in Appendix~\ref{sec:rl-hyperparam}.  
All the experiments can be conducted on a single GPU with 24GB memory.

\begin{table*}[t]
\centering
\begin{tabular}{@{}l|cc|cc|cc@{}}
\toprule
 & \multicolumn{2}{c|}{\textbf{FB15k-237}} & \multicolumn{2}{c|}{\textbf{WN18RR}} & \multicolumn{2}{c}{\textbf{DBpedia50}}  \\
 & Jaccard & Smatch & Jaccard & Smatch & Jaccard & Smatch  \\ \midrule
Encoder-Decoder & 0.565 & \textbf{0.602} & 0.708 & \textbf{0.558} & 0.627 & 0.486 \\
+ RLF-KG (Jaccard) & \textbf{0.666} & 0.530 & 0.753 & 0.540 & \textbf{0.731} & \textbf{0.541} \\
+ RLF-KG (Jaccard + Smatch) & 0.660 & 0.568 & \textbf{0.757} & 0.545 & 0.696 & 0.532  \\ \midrule
Decoder-Only & 0.601 & \textbf{0.614} & 0.692 & \textbf{0.564} & 0.623 & \textbf{0.510}  \\
+ RLF-KG (Jaccard) & \textbf{0.660} & 0.598 & \textbf{0.726} & 0.518 & 0.643 & 0.492 \\
+ RLF-KG (Jaccard + Smatch) & 0.656 & 0.612 & 0.713 & 0.540 & \textbf{0.645} & 0.504  \\ \bottomrule
\end{tabular}
\vspace{-0.1cm}
\caption{The Jaccard and Smatch performance of different reward functions.}
\label{tab:SMATCH_reward}
\end{table*}
\begin{table*}[!htbp]
\centering
\small

\begin{tabular}{@{}l|p{1.45cm}p{0.8cm}p{0.9cm}|p{1.3cm}p{0.8cm}p{0.9cm}|p{1.3cm}p{0.8cm}p{0.9cm}@{}}
\toprule
\textbf{Method} & \multicolumn{3}{c|}{\textbf{FB15k-237}} & \multicolumn{3}{c|}{\textbf{WN18RR}} & \multicolumn{3}{c}{\textbf{DBpedia50}} \\
\multicolumn{1}{c|}{} & Runtime & Jaccard & Smatch & Runtime & Jaccard & Smatch  & Runtime & Jaccard & Smatch \\ \midrule
 Brute-force Search & 16345 mins & 0.635 & 0.305 & 4084 mins & 0.742 & 0.322 & 1132 mins & 0.702 & 0.322 \\
Generation + RLF-KG & \textbf{264 mins} & \textbf{0.666} & \textbf{0.530} & \textbf{32 mins} & \textbf{0.753}& \textbf{0.540} & \textbf{5 mins} & \textbf{0.731} & \textbf{0.541} \\
\bottomrule
\end{tabular}
\vspace{-0.1cm}
\caption{Performance on testing data and runtime for inference for various methods on testing data.}
\label{tab:search-performance}
\end{table*}

\subsection{Experiment Results and Discussions}
\label{sec:exp-result}

We validate RLF-KG effectiveness by comparing the Jaccard metric of the model before and after this process. 
Table~\ref{tab:general_results} displays performance across thirteen hypothesis types on FB15k-237, WN18RR, and DBpedia50. It illustrates Jaccard indices between observations and conclusions of generated hypotheses from the test graph. 
The encoder-decoder and the decoder-only transformers are assessed under fully supervised training on each dataset.
Additionally, performance is reported when models collaborate with reinforcement learning from knowledge graph feedback (RLF-KG).

\subsubsection{Performance Gain after RLF-KG}

We notice  RLF-KG consistently enhances hypothesis generation across three datasets, improving both encoder-decoder and decoder-only models. 
This can be explained by RLF-KG's ability to incorporate knowledge graph information into the generation model, diverging from simply generating hypotheses akin to reference hypotheses.

Additionally, after the RLF-KG training, the encoder-decoder model surpasses the decoder-only structured transformer model.
This is due to the task's nature, where generating a sequence of tokens from an observation set does not necessitate the order of the observation set.
Figure~\ref{fig:reward_figures} supplements the previous statement by illustrating the increasing reward throughout the PPO process.
We also refer readers to Appendix~\ref{sec:case-studies} for qualitative examples demonstrating the improvement in the generated hypotheses for the same observation.

\subsubsection{Adding Structural Reward to PPO}

We explore the potential benefits of incorporating structural similarity into the reward function used in PPO training. 
While RLF-KG originally relies on the Jaccard index, we consider adding the Smatch score, a measure of structural differences between generated and sampled hypotheses. 
We conducted additional experiments to include $S(H, H_{\text{ref}})$ as an extra term in the reward function. 

The results, presented in Table~\ref{tab:SMATCH_reward}, indicate that by incorporating the structural reward, the model can indeed generate hypotheses that are closer to the reference hypotheses {yc in structural sense}. However, the Jaccard scores reveal that with the inclusion of structural information, the overall performance is comparable to or slightly worse than the original reward function. { This is because adding Smatch score tends to cause the generation model to fit or potentially overfit the training data according to the above graph-level similarity, limiting its ability to generalize to unseen test graphs. The Jaccard reward (Eq.~\ref{eq:reward}), however, captures a measure closer to the goal of our task .} Detailed Smatch scores by query types can be found in Appendix~\ref{sec:detailed-smatch-types}.

\subsubsection{Comparison Between Search Methods}

{ We chose the relatively simple brute-force search (Algorithm~\ref{alg:1p-search}) as our search-based baseline due to the inherent high complexity of search algorithms for this task. 
For each observation, the algorithm explores all potential 1p hypotheses within the training graph and selects the one with the highest Jaccard similarity on the training graph.
Despite its simplicity and complexity linear to the number of edges, the algorithm requires significantly more time than our method, not to mention other more complex heuristics.}

{ Following this choice,} we compare inference time and performance between the generation-based and the brute-force search algorithm. Table~\ref{tab:search-performance} { highlights the unsuitability of the brute-force search for scaling up due to its high complexity.} In contrast, generation-based methods demonstrate substantially faster performance. 

Moreover, the generation methods not only outperform the search-based method in Jaccard performance but also show a significant improvement in Smatch performance. { The relatively high Jaccard score of the brute-force search is attributed to its inherent access to an approximation of Jaccard during the search process. Due to the nature of the graph splits, the approximation of Jaccard on the test graph using the training graph for 1p queries is found to be quite accurate, which enhances the average score. The detailed Jaccard scores are presented in Appendix~\ref{sec:search_results}. However, the brute-force search struggles with more complex types of queries.}

\section{Related Work}
The problem of abductive knowledge graph reasoning shares connections with various other knowledge graph reasoning tasks, including knowledge graph completion, complex logical query answering, and rule mining. 

\paragraph{Rule Mining} Rule mining focuses on inductive logical reasoning, namely discovering logical rules over the knowledge graph. 
Various methods are proposed in this line of work \citep{DBLP:journals/vldb/GalarragaTHS15,DBLP:conf/semweb/HoSGKW18,DBLP:conf/ijcai/MeilickeCRS19, DBLP:conf/kdd/ChengL0S22, DBLP:conf/iclr/ChengAS23}. { While the direct application of rule mining to solve this task is theoretically applicable, rule mining algorithms such as those described by \citet{DBLP:journals/vldb/GalarragaTHS15} also rely on searches to construct the horn clauses, which face scalability issues.}

\paragraph{Complex logical query} Complex logical query answering is a task of answering logically structured queries on KG.  
Query embedding primary focus is the enhancement of embedding structures for encoding sets of answers \citep{DBLP:conf/nips/HamiltonBZJL18, sun2020faithful,liu2021neural, DBLP:conf/kdd/BaiLLYYS23,DBLP:conf/nips/BaiLW0S23,DBLP:journals/corr/abs-2312-13866,DBLP:journals/corr/abs-2312-15591,DBLP:journals/corr/abs-2307-13701,DBLP:conf/acl/WangFYSWS23,DBLP:conf/iclr/WangSWS23,DBLP:journals/corr/abs-2402-14609,DBLP:conf/www/LiuWBST24}. For instance, \citet{DBLP:conf/nips/RenL20} and \citet{zhang2021cone} introduce the utilization of geometric structures such as rectangles and cones within hyperspace to represent entities. 
Neural MLP (Mixer) \citep{AmayuelasZR022} use MLP and MLP-Mixer as the operators. 
\cite{bai-etal-2022-query2particles} suggests employing multiple vectors to encode queries, thereby addressing the diversity of answer entities.
 FuzzQE \citep{chen2022fuzzy} uses fuzzy logic to represent logical operators.
Probabilistic distributions can also serve as a means of query encoding \citep{choudhary2021probabilistic, choudhary2021self}, with examples including Beta Embedding \citep{DBLP:conf/nips/RenL20} and Gamma Embedding \citep{yang2022gammae}.

\section{Conclusion}
In summary, this paper has introduced the task of abductive logical knowledge graph reasoning. Meanwhile, this paper has proposed a generation-based method to address knowledge graph incompleteness and reasoning efficiency by generating logical hypotheses. Furthermore, this paper demonstrates the effectiveness of 
our proposed reinforcement learning from knowledge graphs (RLF-KG) to enhance our hypothesis generation model by leveraging feedback from knowledge graphs.

\section*{Limitations}

Our proposed methods and techniques in the paper are evaluated on a specific set of knowledge graphs, namely FB15k-237, WN18RR, and DBpedia50. It is unclear how well these approaches would perform on other KGs with different characteristics or domains.
Meanwhile, knowledge graphs can be massive and continuously evolving, our method is not yet able to address the dynamic nature of knowledge evolutions, like conducting knowledge editing automatically. It is important to note that these limitations should not undermine the significance of the work but rather serve as areas for future research and improvement.

\section*{Acknowledgements}
We thank the anonymous reviewers and the area chair for their constructive comments.
The authors of this paper were supported by the NSFC Fund (U20B2053) from the NSFC of China, the RIF (R6020-19 and R6021-20), and the GRF (16211520 and 16205322) from RGC of Hong Kong. 
We also thank the support from the UGC Research Matching Grants (RMGS20EG01-D, RMGS20CR11, RMGS20CR12, RMGS20EG19, RMGS20EG21, RMGS23CR05, RMGS23EG08). 
\bibliography{anthology,custom}
\bibliographystyle{acl_natbib}

\appendix

\newpage

\section{Clarification on Abductive, Deductive, and Inductive Reasoning}
\label{app:clar-abd-ded-ind}

Here we use simple syllogisms to explain the connections and differences between abductive reasoning and the other two types of reasoning, namely, deductive and inductive reasoning.
In deductive reasoning, the inferred conclusion is necessarily true if the premises are true. Suppose we have a major premise $P_1$: \textit{All men are mortal}, a minor premise $P_2$: \textit{Socrates is a man}, then we can conclude $C$ that \textit{Socrates is mortal}. 
This can also be expressed as the inference $ \infer{C}{P_1 \land P_2}$.
On the other hand, abductive reasoning aims to explain an observation and is non-necessary, i.e., the inferred hypothesis is not guaranteed to be true. We also start with a premise $P$:  \textit{All cats like catching mice}, and then we have some observation $O$: \textit{Katty like catching mice}. 
The abduction gives a simple yet most probable hypothesis \(H\): \textit{Katty is a cat}, as an explanation. This can also be written as the inference \(\infer{H}{P \land O}\). Different than deductive reasoning, the observation $O$ should be entailed by the premise $P$ and the hypotheses $H$, which can be expressed by the implication $P \land H \implies O$. 
The other type of non-necessary reasoning is inductive reasoning, where, in contrast to the appeal to explanatory considerations in abductive reasoning, there is an appeal to observed frequencies \citep{abduction}. For instance, premises \(P_1\): \textit{Most Math students learn linear algebra in their first year} and \(P_2\): \text{Alice is a Math student} infer \(H\): \textit{Alice learned linear algebra in her first year}, i.e., \(\infer{H}{P_1 \land P_2}\). Note that the inference rules in the last two examples are not strictly logical implications.

It is worth mentioning that there might be different definitions or interpretations of these forms of reasoning.

\section{Example of Observation-Hypothesis Pair}
\label{app:obse-hypo-eg}
For example, observation $O$ can be a set of name entities like $\{ Grant Heslov$, $Jason Segel$, $Robert Towne$, $Ronald Bass, Rashida Jones\}$.
Given this observation, an abductive reasoner is required to give the logical hypothesis that best explains it. 
For the above example, the expected hypothesis \(H\) in natural language is that they are \textit{actors} and \textit{screenwriters}, and they are also born in \textit{Los Angeles}. Mathematically, the hypothesis \(H\) can be represented by a logical expression of the facts of the KG: 
$H(V) = Occupation(V, Actor)$ $\land$ $Occupation(V, ScreenWriter)$ $\land$ $BornIn(V, Los Angeles)$.
Although the logical expression here only contains logical conjunction \texttt{AND} ($\land$), we consider more general first-order logical form as defined in Section~\ref{sec:prob-form}.

\section{Statistics of Queries Sampled for Datasets}
\label{sec:queries-details}

Table~\ref{tab:queries_details} presents the numbers of queries sampled for each dataset in each stage.

\begin{table*}[htbp]
\begin{center}

\begin{tabular}{l|cc|cc|cc}
\toprule

\multirow{2}{*}{\textbf{Dataset}} & \multicolumn{2}{c|}{\textbf{Training Samples}} & \multicolumn{2}{c|}{\textbf{Validation Samples}} & \multicolumn{2}{c}{\textbf{Testing Samples}} \\
       & Each Type & Total  & Each Type & Total & Each Type & Total \\ \midrule
FB15k-237   & 496,126 & 6,449,638 & 62,015 & 806,195 & 62,015 & 806,195  \\
WN18RR   &  148,132  & 1,925,716  &  18,516   & 240,708   & 18,516 & 240,708\\
DBpedia50   &  55,028  & 715,364  &  6,878   & 89,414  &  6,878 &  89,414 \\
\bottomrule

\end{tabular}
\caption{The detailed information about the queries used for training, validation, and testing. }
\label{tab:queries_details}
\end{center}
\vspace{-0.5cm}
\end{table*}

\section{Algorithm for Sampling Observation-Hypothesis Pairs}
\label{sec:algo-sampling}

Algorithm \ref{alg:grounding_queries} is designed for sampling complex hypotheses from a given knowledge graph. Given a knowledge graph $\gG$ and a hypothesis type $T$, the algorithm starts with a random node $v$ and proceeds to recursively construct a hypothesis that has $v$ as one of its conclusions and adheres the type $T$. 
During the recursion process, the algorithm examines the last operation in the current hypothesis. If the operation is a \textit{projection}, the algorithm randomly selects one of \(v\)'s in-edge $(u, r, v)$.
Then, the algorithm calls the recursion on node $u$ and the sub-hypothesis type of $T$ again. If the operation is an \textit{intersection}, it applies recursion on the sub-hypotheses and the same node $v$. If the operation is a \textit{union}, it applies recursion on one sub-hypothesis with node \(v\) and on other sub-hypotheses with an arbitrary node, as \text{union} only requires one of the sub-hypotheses to have \(v\) as an answer node. 
The recursion stops when the current node contains an entity.

\begin{algorithm}
	\caption{Sampling Hypothesis According to Type} 
	\label{alg:grounding_queries}
    \hspace*{\algorithmicindent} \textbf{Input} Knowledge graph $\gG $, hypothesis type $T$\\
    \hspace*{\algorithmicindent} \textbf{Output} Hypothesis sample
	\begin{algorithmic}[t]    
	    \Function {GroundType}{$\gG, T, t$}
    	    \If {$T.operation = p$}
    	        \State $(h, r) \gets \Call{Sample} {\{(h, r)| (h, r, t) \in \gE(\gG) \}} $
                \State $\hat{T} \gets $ the only subtype in \(T.SubTypes\)
    	        \State $H \gets \Call{GroundType}{\gG, \hat{T}, h}$  
    	        \State \Return $(p, r, H)$
    	    \ElsIf{$T.operation = i$ }
    	        \State {$SubHypotheses \gets \emptyset $}
    	        \For {$\hat{T}  \in T.SubTypes$}
    	            \State $H \gets\Call{GroundType}{\gG, \hat{T}, t}$
    	            \State {$SubHypotheses.\Call{pushback}{H} $}
    	        \EndFor
    	        \State \Return $(i, SubHypotheses)$
    	   \ElsIf{$T.operation = u$ }
    	        \State {$SubHypotheses \gets \emptyset $}
    	        \For {$\hat{T}  \in T.SubTypes$}
        	        \If{$\hat{T}$ is the first subtype} 
                        \State $H \gets\Call{GroundType}{\gG, \hat{T}, t}$        	           
    	            \Else
                        \State {$\hat{t} \gets \Call{Sample}{\gV(\gG)}$} 
    	               \State $H \gets\Call{GroundType}{\gG, \hat{T}, \hat{t}}$
    	            \EndIf
    	            \State {$SubHypotheses.\Call{pushback}{H} $}
    	        \EndFor
    	        \State \Return $(u, SubHypotheses)$
    	    \ElsIf{$T.operation = e$ }
    	            \State \Return $(e, t)$
    	    \EndIf
	    \EndFunction
    \State $v \gets \Call{Sample}{\gV(\gG)}$
    \State \Return \Call{GroundType}{$\gG, T, v$}
	\end{algorithmic} 
\end{algorithm}

\section{Algorithms for Conversion between Queries and Actions}
\label{sec:algo-queries-actions}

We here present the details of tokenizing the hypothesis graph (Algorithm \ref{alg:q2a}), and formulating a graph according to the tokens, namely the process of de-tokenization (Algorithm \ref{alg:a2q}).
Inspired by the action-based semantic parsing algorithms, we view tokens as actions.
It is worth noting that we employ the symbols \(G, V, E\) for the hypothesis graph to differentiate it from the knowledge graph.

\begin{algorithm}[!htbp]
\caption{HypothesisToActions}
\label{alg:q2a}
\hspace*{\algorithmicindent} \textbf{Input} Hypothesis plan graph $G$\\
\hspace*{\algorithmicindent} \textbf{Output} Action sequence $A$
\begin{algorithmic}
\Function{DFS}{$G, t, A$}
\If{$t$ is an anchor node}
\State \(action \gets \) entity associated with \(t\)
\Else
\State \(action \gets \) operator associated with the first in-edge of \(t\)
\EndIf
\State  \(A.\Call{PushBack}{action}\)
\ForAll{in-edges to $t$ in $G$ $(h, r, t)$}
\State \Call{DFS}{$G, h, A$}
\EndFor
\EndFunction
\State \(root \gets \) the root of \(G\)
\State \(A \gets \) \Call{DFS}{$G, root, \emptyset$}
\State \Return \(A\)
\end{algorithmic}
\end{algorithm}

\begin{algorithm}[!htbp]
\caption{ActionsToHypothesis}
\label{alg:a2q}
\hspace*{\algorithmicindent} \textbf{Input} Action sequence $A$  \\
\hspace*{\algorithmicindent} \textbf{Output} Hypothesis plan graph $G$
\begin{algorithmic}
\State \(S \gets \) an empty stack
\State Create an map \(deg\). \(deg[i] = deg[u] = 2\) and \(=1\) otherwise.
\State \(V \gets \emptyset, E \gets \emptyset\)
\For{$a \in A$}
\State Create a new node \(h\), \(V \gets V \cup \{h\}\)
\If{\(S \neq \emptyset\)}
\State \((t, operator, d) \gets S.\Call{Top}{}\)
\State \(E \gets E \cup \{(h, operator, t)\}\)
\EndIf
\If{\(a\) represents an anchor}
\State Mark \(h\) as an anchor with entity \(a\)
\While{\(S \neq \emptyset\)}
\State \((t, operator, d) \gets S.\Call{Pop}{}\) 
\State \(d \gets d - 1\)
\If{\(d > 0\)}
\State $S.\Call{PushBack}{(t, operator, d)}$
\State Break
\EndIf
\EndWhile
\Else
\State $S.\Call{PushBack}{(h, a, deg[a])}$
\EndIf
\EndFor
\State \(G \gets (V, E)\)
\State \Return \(G\)
\end{algorithmic}
\end{algorithm}

\section{Details of using Smatch to evaluate structural differneces of queries}
\label{sec:Smatch}
Smatch \cite{DBLP:conf/acl/CaiK13} is an evaluation metric for Abstract Meaning Representation (AMR) graphs. 
An AMR graph is a directed acyclic graph with two types of nodes: variable nodes and concept nodes, and three types of edges:
\begin{itemize}
\item Instance edges, which connect a variable node to a concept node and are labeled literally ``instance''. Every variable node must have exactly one instance edge, and vice versa.
\item Attribute edges, which connect a variable node to a concept node and are labeled with attribute names.
\item Relation edges, which connect a variable node to another variable node and are labeled with relation names.
\end{itemize}
Given a predicted AMR graph \(G_{\text{p}}\) and the gold AMR graph \(G_{\text{g}}\), the Smatch score of \(G_{\text{p}}\) with respect to \(G_{\text{g}}\) is denoted by \(\mathtt{Smatch}(G_{\text{p}}, G_{\text{g}})\). \(\mathtt{Smatch}(G_{\text{p}}, G_{\text{g}})\) is obtained by finding an approximately optimal mapping between the variable nodes of the two graphs and then matching the edges of the graphs. 

Our hypothesis graph is similar to the AMR graph, in:
\begin{itemize}
\item The nodes are both categorized as fixed nodes and variable nodes
\item The edges can be categorized into two types: edges from a variable node to a fixed node and edges from a variable node to another variable node. And edges are labeled with names.
\end{itemize}
However, they are different in that the AMR graph requires every variable node to have instance edges, while the hypothesis graph does not.

The workaround for leveraging the Smatch score to measure the similarity between hypothesis graphs is creating an instance edge from every entity to some virtual node. Formally, given a hypothesis \(H\) with hypothesis graph \(G(H)\), we create a new hypothesis graph \(G_A(H)\) to accommodate the AMR settings as follows: First, we initialize \(G_A(H) = G(H)\). Then, create a new relation type \(instance\) and add a virtual node \(v'\) into \(G_A(H)\). Finally, for every variable node \(v \in G(H)\), we add a relation \(instance(v, v')\) into \(G_{A}(H)\). Then, given a predicted hypothesis \(H_{p}\) and a gold hypothesis \(H_{g}\), the Smatch score is defined as
\begin{equation}
    S(H_{p}, H_{g}) = \mathtt{Smatch}(G_A(H_{p}), G_A(H_{g})).
\end{equation}

Through this conversion, a variable entity \(v_{g}\) of \(H_{\text{g}}\) is mapped to a variable entity \(v_{p}\) of \(H_{\text{p}}\) if and only if \(instance(v_{g}, v')\) is matched with \(instance(v_{p}, v')\). This modification does not affect the overall algorithm for finding the optimal mapping between variable nodes and hence gives a valid and consistent similarity score. However, this adds an extra point for matching between instance edges, no matter how the variable nodes are mapped.

\section{Detailed Smatch Scores by Query Types}
\label{sec:detailed-smatch-types}

Tables~\ref{tab:smatch} and \ref{tab:search-s} show the detailed Smatch performance of various methods.

\begin{table*}[!htbp]
\centering
\tiny
\begin{tabular}
{@{}l|l|p{0.3cm}p{0.3cm}p{0.3cm}p{0.3cm}p{0.3cm}p{0.3cm}p{0.3cm}p{0.4cm}|p{0.3cm}p{0.3cm}p{0.3cm}p{0.3cm}p{0.4cm}|c@{}}
\toprule
\textbf{Dataset} & \textbf{Model} & \textbf{1p} & \textbf{2p} & \textbf{2i} & \textbf{3i} & \textbf{ip} & \textbf{pi} & \textbf{2u} & \textbf{up} & \textbf{2in} & \textbf{3in} & \textbf{pni} & \textbf{pin} & \textbf{inp} & \textbf{Ave.} \\ \midrule
\multirow{6}{*}{FB15k-237} & Enc.-Dec. & 0.342 & 0.506 & 0.595 & 0.602 & 0.570 & 0.598 & 0.850 & 0.571 & 0.652 & 0.641 & 0.650 & 0.655 & 0.599 & 0.602 \\
 & RLF-KG (J) & 0.721 & 0.643 & 0.562 & 0.480 & 0.364 & 0.475 & 0.769 & 0.431 & 0.543 & 0.499 & 0.513 & 0.518 & 0.370 & 0.530 \\
 & RLF-KG (J+S) & 0.591 & 0.583 & 0.577 & 0.531 & 0.447 & 0.520 & 0.820 & 0.505 & 0.602 & 0.563 & 0.571 & 0.595 & 0.484 & 0.568 \\\cmidrule(l){2-16} 
 & Dec.-Only & 0.287 & 0.481 & 0.615 & 0.623 & 0.599 & 0.626 & 0.847 & 0.574 & 0.680 & 0.656 & 0.675 & 0.677 & 0.636 & 0.614 \\
 & RLF-KG (J) & 0.344 & 0.445 & 0.675 & 0.585 & 0.537 & 0.638 & 0.853 & 0.512 & 0.696 & 0.575 & 0.647 & 0.688 & 0.574 & 0.598 \\
 & RLF-KG (J+S) & 0.303 & 0.380 & 0.692 & 0.607 & 0.565 & 0.671 & 0.857 & 0.506 & 0.727 & 0.600 & 0.676 & 0.734 & 0.634 & 0.612 \\ \midrule
\multirow{6}{*}{WN18RR} & Enc.-Dec. & 0.375 & 0.452 & 0.591 & 0.555 & 0.437 & 0.585 & 0.835 & 0.685 & 0.586 & 0.516 & 0.561 & 0.549 & 0.530 & 0.558 \\
 & RLF-KG (J) & 0.455 & 0.468 & 0.563 & 0.562 & 0.361 & 0.530 & 0.810 & 0.646 & 0.560 & 0.530 & 0.536 & 0.539 & 0.465 & 0.540 \\
 & RLF-KG (J+S) & 0.443 & 0.457 & 0.565 & 0.572 & 0.366 & 0.545 & 0.814 & 0.661 & 0.541 & 0.553 & 0.532 & 0.546 & 0.491 & 0.545 \\ \cmidrule(l){2-16} 
 & Dec.-Only & 0.320 & 0.443 & 0.582 & 0.551 & 0.486 & 0.597 & 0.809 & 0.696 & 0.594 & 0.526 & 0.575 & 0.574 & 0.577 & 0.564 \\
 & RLF-KG (J) & 0.400 & 0.438 & 0.566 & 0.491 & 0.403 & 0.519 & 0.839 & 0.667 & 0.547 & 0.450 & 0.497 & 0.466 & 0.450 & 0.518 \\
 & RLF-KG (J+S) & 0.375 & 0.447 & 0.584 & 0.499 & 0.432 & 0.545 & 0.825 & 0.679 & 0.584 & 0.477 & 0.543 & 0.522 & 0.507 & 0.540 \\ \midrule
\multirow{6}{*}{DBpedia50} & Enc.-Dec. & 0.345 & 0.396 & 0.570 & 0.548 & 0.344 & 0.576 & 0.712 & 0.544 & 0.474 & 0.422 & 0.477 & 0.488 & 0.428 & 0.486 \\
 & RLF-KG (J) & 0.461 & 0.424 & 0.634 & 0.584 & 0.361 & 0.575 & 0.809 & 0.579 & 0.584 & 0.497 & 0.544 & 0.533 & 0.454 & 0.541 \\
 & RLF-KG (J+S) & 0.419 & 0.410 & 0.638 & 0.555 & 0.373 & 0.586 & 0.785 & 0.579 & 0.560 & 0.459 & 0.536 & 0.542 & 0.474 & 0.532 \\ \cmidrule(l){2-16} 
 & Dec.-Only & 0.378 & 0.408 & 0.559 & 0.526 & 0.397 & 0.568 & 0.812 & 0.626 & 0.480 & 0.414 & 0.489 & 0.494 & 0.474 & 0.510 \\
 & RLF-KG (J) & 0.405 & 0.411 & 0.558 & 0.496 & 0.376 & 0.507 & 0.825 & 0.621 & 0.477 & 0.397 & 0.468 & 0.444 & 0.406 & 0.492 \\
 & RLF-KG (J+S) & 0.398 & 0.415 & 0.567 & 0.497 & 0.383 & 0.533 & 0.827 & 0.630 & 0.510 & 0.420 & 0.484 & 0.457 & 0.430 & 0.504 \\ \bottomrule
\end{tabular}
\caption{The detailed Smatch performance of various methods.}
\label{tab:smatch}
\end{table*}

\begin{table*}[!htbp]
\centering
\small
\begin{tabular}{@{}l|p{0.45cm}p{0.45cm}p{0.45cm}p{0.45cm}p{0.45cm}p{0.45cm}p{0.45cm}p{0.60cm}|p{0.45cm}p{0.45cm}p{0.45cm}p{0.45cm}p{0.60cm}|c@{}}
\toprule
\textbf{Dataset} & \textbf{1p} & \textbf{2p} & \textbf{2i} & \textbf{3i} & \textbf{ip} & \textbf{pi} & \textbf{2u} & \textbf{up} & \textbf{2in} & \textbf{3in} & \textbf{pni} & \textbf{pin} & \textbf{inp} & \textbf{Ave.} \\ \midrule
FB15k-237 & 0.945 & 0.340 & 0.365 & 0.218 & 0.184 & 0.267 & 0.419 & 0.185 & 0.301 & 0.182 & 0.245 & 0.155 & 0.157 & 0.305 \\
WN18RR & 0.957 & 0.336 & 0.420 & 0.274 & 0.182 & 0.275 & 0.427 & 0.183 & 0.323 & 0.224 & 0.270 & 0.155 & 0.156 & 0.322 \\
DBpedia & 0.991 & 0.336 & 0.399 & 0.259 & 0.182 & 0.245 & 0.441 & 0.183 & 0.332 & 0.226 & 0.290 & 0.154 & 0.155 & 0.322 \\ \bottomrule
\end{tabular}
\caption{The detailed Smatch performance of the searching method.}
\label{tab:search-s}
\end{table*}

\section{Detailed Jaccard Performance of the Brute-force Search}
\label{sec:search_results}

Table~\ref{tab:search_results} shows the detailed Jaccard performance of the brute-force search.

\begin{table*}[!ht]
    \tiny
    \centering
    \begin{tabular}{l|l|l|l|l|l|l|l|l|l|l|l|l|l|l}
    \toprule
        Dataset  & 1p  & 2p  & 2i  & 3i  & ip  & pi  & 2u  & up  & 2in  & 3in  & pni  & pin  & inp  & Ave. \\ \midrule
        FB15k-237  & 0.980  & 0.563  & 0.639  & 0.563  & 0.732  & 0.633  & 0.744  & 0.585  & 0.659  & 0.479  & 0.607  & 0.464  & 0.603  & 0.635 \\ 
        WN18RR  & 0.997  & 0.622  & 0.784  & 0.776  & 0.955  & 0.666  & 0.753  & 0.605  & 0.783  & 0.757  & 0.762  & 0.540  & 0.630  & 0.741  \\ 
        DBpedia  & 0.997  & 0.705  & 0.517  & 0.517  & 0.982  & 0.461  & 0.783  & 0.754  & 0.722  & 0.658  & 0.782  & 0.544  & 0.700  & 0.702 \\ \bottomrule
    \end{tabular}
    \caption{The detailed Jaccard performance of the brute-force search.}
    \label{tab:search_results}
\end{table*}

\section{Hyperparameters of the RL Process}
\label{sec:rl-hyperparam}
The PPO hyperparameters are shown in Table~\ref{tab:ppo-hyperparam}.
We warm-uped the learning rate from \(0.1\) of the peak to the peak value within the first \(10\%\) of total iterations, followed by a decay to \(0.1\) of the peak using a cosine schedule.
\begin{table*}[!htbp]
\centering
\begin{tabular}{@{}l|lll|lll@{}}
\toprule
\multirow{2}{*}{\textbf{Hyperparam.}} 
& \multicolumn{3}{c|}{\textbf{Enc.-Dec.}} & \multicolumn{3}{c}{\textbf{Dec.-Only}} \\ 
 & FB15k-237 & WN18RR & DBpedia50 & FB15k-237 & WN18RR & DBpedia50 \\ \midrule
Learning rate & 2.4e-5 & 3.1e-5 & 1.8e-5 & 0.8e-5 & 0.8e-5 & 0.6e-5 \\
Batch size & 16384 & 16384 & 4096 & 3072 & 2048 & 2048 \\
Minibatch size & 512 & 512 & 64 & 96 & 128 & 128 \\
Horizon & 4096 & 4096 & 4096 & 2048 & 2048 & 2048 \\ \bottomrule
\end{tabular}
\caption{PPO Hyperparamters.}
\label{tab:ppo-hyperparam}
\end{table*}

\section{Algorithms for One-Hop Searching}
\label{sec:algo-search}
We now introduce the Algorithm~\ref{alg:1p-search} used for searching the best one-hop hypothesis with the tail among all entities in the observation set to explain the observations.

\begin{algorithm}[!htbp]
\caption{One-Hop-Search}
\label{alg:1p-search}
\hspace*{\algorithmicindent} \textbf{Input} Observation set \(O\)\\
\hspace*{\algorithmicindent} \textbf{Output} Hypothesis $bestHypothesis$
\begin{algorithmic}
\State \(candidates \gets \{(h, r, t) \in \gR_{\text{train}} | t \in O \}\)
\State \(bestJaccard \gets 0\)
\State \(bestHypothesis \gets\) Null
\For{\((h,r,t) \in candidates\)}
\State \(H \gets\) the one-hop hypothesis consisting of edge \((h,r,t)\)
\State \(nowJaccard \gets \mathtt{Jaccard}(\ans{H}_{\gG_{\text{train}}}, A)\)
\If{\(nowJaccard > bestJaccard\)}
\State \(bestJaccard \gets nowJaccard\)
\State \(bestHypothesis \gets H\)
\EndIf
\EndFor
\State \Return \(bestHypothesis\)
\end{algorithmic}
\end{algorithm}

\section{Case Studies}
\label{sec:case-studies}

Explore Table~\ref{tab:fb-case1}, \ref{tab:fb-case2} and \ref{tab:db-case1} for concrete examples generated by various abductive reasoning methods, namely search, generative model with supervised training, and generative model with supervised training incorporating RLF-KG. 

\begin{table*}[t]
\centering
\small
\begin{tabular}{@{}l|l|p{5cm}l@{}}
\toprule
\multirow{16}{*}{\textbf{Sample}} & Interpretation & \multicolumn{2}{p{10cm}}{\begin{tabular}[c]{@{}l@{}}Companies operating in industries that intersect with  Yahoo! but not with IBM.\end{tabular}} \\ \cmidrule(l){2-4} 
 & Hypothesis & \multicolumn{2}{p{10cm}}{The observations are the $V_?$ such that $\exists V_1, inIndustry(V_1, V_?) \land \neg industryOf(IBM, V_1) \land industryOf(Yahoo!, V_1)$} \\ \cmidrule(l){2-4} 
 & \multirow{14}{*}{Observation} & EMI, & CBS\_Corporation, \\
 &  & Columbia, & GMA\_Network, \\
 &  & Viacom, & Victor\_Entertainment, \\
 &  & Yahoo!, & Sony\_Music\_Entertainment\_(Japan)\_Inc., \\
 &  & Bandai, & Toho\_Co.,\_Ltd., \\
 &  & Rank\_Organisation, & The\_New\_York\_Times\_Company, \\
 &  & Gannett\_Company, & Star\_Cinema, \\
 &  & NBCUniversal, & TV5, \\
 &  & Pony\_Canyon, & Avex\_Trax, \\
 &  & The\_Graham\_Holdings\_Company, & The\_Walt\_Disney\_Company, \\
 &  & Televisa, & Metro-Goldwyn-Mayer, \\
 &  & Google, & Time\_Warner, \\
 &  & Microsoft\_Corporation, & Dell, \\
 &  & Munhwa\_Broadcasting\_Corporation, & News\_Corporation \\ \midrule
\multirow{5}{*}{\textbf{Searching}} & Interpretation & \multicolumn{2}{p{10cm}}{Which companies operate in media industry?} \\ \cmidrule(l){2-4} 
 & Hypothesis & \multicolumn{2}{p{10cm}}{The observations are the $V_?$ such that $inIndustry(Media, V_?)$} \\ \cmidrule(l){2-4} 
 & Conclusion & \begin{tabular}[c]{@{}l@{}}Absent:\\ - Google,\\ - Microsoft\_Corporation,\\ - Dell\end{tabular} &  \\ \cmidrule(l){2-4} 
 & Jaccard & 0.893 &  \\ \cmidrule(l){2-4} 
 & Smatch & 0.154 &  \\ \midrule
\multirow{5}{*}{\textbf{Enc.-Dec.}} & Interpretation & \multicolumn{2}{p{10cm}}{\begin{tabular}[c]{@{}l@{}}Companies operating in industries that intersect with \\ Yahoo! but not with Microsoft Corporation.\end{tabular}} \\ \cmidrule(l){2-4} 
 & Hypothesis & \multicolumn{2}{p{10cm}}{The observations are the $V_?$ such that $\exists V_1, inIndustry(V_1, V_?) \land \neg industryOf(Microsoft\_Corporation, V_1) \land industryOf(Yahoo!, V_1)$} \\ \cmidrule(l){2-4} 
 & Conclusion & Absent: Microsoft\_Corporation &  \\ \cmidrule(l){2-4} 
 & Jaccard & 0.964 &  \\ \cmidrule(l){2-4} 
 & Smatch & 0.909 &  \\ \midrule
\multirow{5}{*}{\textbf{+ RLF-KG}} & Interpretation & \begin{tabular}[c]{@{}l@{}}Companies operating in industries that intersect with\\ Yahoo! but not with Oracle Corporation.\end{tabular} &  \\ \cmidrule(l){2-4} 
 & Hypothesis & \multicolumn{2}{p{10cm}}{The observations are the $V_?$ such that $\exists V_1, inIndustry(V_1, V_?) \land \neg industryOf(Oracle\_Corporation, V_1) \land industryOf(Yahoo!, V_1)$} \\ \cmidrule(l){2-4} 
 & Concl. & Same &  \\ \cmidrule(l){2-4} 
 & Jaccard & 1.000 &  \\ \cmidrule(l){2-4} 
 & Smatch & 0.909 &  \\ \bottomrule
\end{tabular}
\caption{Example FB15k-237 Case study 1.}
\label{tab:fb-case1}
\end{table*}

\begin{table*}[t]
\centering
\small
\begin{tabular}{@{}l|l|p{5cm}l@{}}
\toprule
\multirow{5}{*}{\textbf{Sample}} & Interpretation & \multicolumn{2}{p{10cm}}{\begin{tabular}[c]{@{}l@{}}Locations that adjoin second-level divisions of the United \\ States of America that adjoin Washtenaw County.\end{tabular}} \\ \cmidrule(l){2-4} 
 & Hypothesis & \multicolumn{2}{p{10cm}}{The observations are the $V_?$ such that $\exists V_1, adjoins(V_1, V_?) \land adjoins(Washtenaw\_County, V_1) \land secondLevelDivisions(USA, V_1)$} \\ \cmidrule(l){2-4} 
 & \multirow{3}{*}{Observation} & Jackson\_County, & Macomb\_County, \\
 &  & Wayne\_County, & Ingham\_County \\
 &  & Washtenaw\_County, &  \\ \midrule
\multirow{5}{*}{\textbf{Searching}} & Interpretation & \multicolumn{2}{p{10cm}}{Locations that adjoin Oakland County.} \\ \cmidrule(l){2-4} 
 & Hypothesis & \multicolumn{2}{p{10cm}}{The observations are the $V_?$ such that $adjoins(Oakland\_County, V_?)$} \\ \cmidrule(l){2-4} 
 & Conclusion & \begin{tabular}[c]{@{}l@{}}Absent:\\ - Jackson\_County\\ - Ingham\_County\end{tabular} &  \\ \cmidrule(l){2-4} 
 & Jaccard & 0.600 &  \\ \cmidrule(l){2-4} 
 & Smatch & 0.182 &  \\ \midrule
\multirow{5}{*}{\textbf{Enc.-Dec.}} & Interpretation & \multicolumn{2}{p{10cm}}{\begin{tabular}[c]{@{}l@{}}Second-level divisions of the United States of America \\ that adjoin locations that adjoin Oakland County.\end{tabular}} \\ \cmidrule(l){2-4} 
 & Hypothesis & \multicolumn{2}{p{10cm}}{The observations are the $V_?$ such that $\exists V_1, secondLevelDivisions(USA, V_?) \land adjoins(V_1, V_?) \land +adjoins(Oakland\_County, V_1)$} \\ \cmidrule(l){2-4} 
 & Conclusion & \begin{tabular}[c]{@{}l@{}}Extra: Oakland\_County\\ Absent: Wayne\_County\end{tabular} &  \\ \cmidrule(l){2-4} 
 & Jaccard & 0.667 &  \\ \cmidrule(l){2-4} 
 & Smatch & 0.778 &  \\ \midrule
\multirow{5}{*}{\textbf{+ RLF-KG}} & Interpretation & \multicolumn{2}{p{10cm}}{\begin{tabular}[c]{@{}l@{}}Second-level divisions of the United States of America \\ that adjoin locations contained within Michigan.\end{tabular}} \\ \cmidrule(l){2-4} 
 & Hypothesis & \multicolumn{2}{p{10cm}}{The observations are the $V_?$ such that $\exists V_1, secondLevelDivisions(USA, V_?) \land adjoins(V_1, V_?) \land containedIn(Michigan, V_1)$} \\ \cmidrule(l){2-4} 
 & Conclusion & \begin{tabular}[c]{@{}l@{}}Extra:\\ - Oakland\_County\\ - Genesee\_County\end{tabular} &  \\ \cmidrule(l){2-4} 
 & Jaccard & 0.714 &  \\ \cmidrule(l){2-4} 
 & Smatch & 0.778 &  \\ \bottomrule
\end{tabular}
\caption{FB15k-237 Case study 2.}
\label{tab:fb-case2}
\end{table*}

\begin{table*}[t]
\centering
\small
\begin{tabular}{@{}l|l|p{5cm}p{3cm}}
\toprule
\multirow{6}{*}{\textbf{Ground Truth}} & Interpretation & \multicolumn{2}{p{10cm}}{\begin{tabular}[c]{@{}l@{}}Works, except for “Here ’Tis,” that have subsequent works in the jazz genre.\end{tabular}} \\ \cmidrule(l){2-4} 
 & Hypothesis & \multicolumn{2}{p{10cm}}{ The observations are the $V_?$ such that $\exists V_1, subsequentWork(V_1, V_?) \land \neg previousWork(Here\_'Tis, V_1) \land genre(Jazz, V_1) $} \\ \cmidrule(l){2-4} 
 & \multirow{4}{*}{Observation} & Deep,\_Deep\_Trouble, & Lee\_Morgan\_Sextet, \\
 &  & Good\_Dog,\_Happy\_Man, & Paris\_Nights\textbackslash{}/New\_York\_Mornings, \\
 &  & I\_Don't\_Want\_to\_Be\_Your\_Friend, & Take\_the\_Box \\
 &  & Interior\_Music, &  \\ \midrule
\multirow{5}{*}{\textbf{Searching}} & Interpretation & \multicolumn{2}{p{10cm}}{Works subsequent to “Closer” (Corinne Bailey Rae song).} \\ \cmidrule(l){2-4} 
 & Hypothesis & \multicolumn{2}{p{10cm}}{The observations are the $V_?$ such that $subsequentWork(Closer\_(Corinne\_Bailey\_Rae\_song), V_?)$} \\ \cmidrule(l){2-4} 
 & Conclusion & Only Paris\_Nights\textbackslash{}/New\_York\_Mornings &  \\ \cmidrule(l){2-4} 
 & Jaccard & 0.143 &  \\ \cmidrule(l){2-4} 
 & Smatch & 0.154 &  \\ \midrule
\multirow{5}{*}{\textbf{Enc.-Dec.}} & Interpretation & \multicolumn{2}{p{10cm}}{\begin{tabular}[c]{@{}l@{}}Works, except for “Lee Morgan Sextet,” that have subsequent works \\ in the jazz genre.\end{tabular}} \\ \cmidrule(l){2-4} 
 & Hypothesis & \multicolumn{2}{p{10cm}}{The observations are the $V_?$ such that $\exists V_1, subsequentWork(V_1, V_?) \land \neg previousWork(Lee\_Morgan\_Sextet, V_1) \land genre(Jazz, V_1)$} \\ \cmidrule(l){2-4} 
 & Conclusion & \begin{tabular}[c]{@{}l@{}}Extra: Here\_'Tis\\ Absent: Lee\_Morgan\_Sextet\end{tabular} &  \\ \cmidrule(l){2-4} 
 & Jaccard & 0.750 &  \\ \cmidrule(l){2-4} 
 & Smatch & 0.909 &  \\ \midrule
\multirow{5}{*}{\textbf{+ RLF-KG}} & Interpretation & \multicolumn{2}{p{10cm}}{Works that have subsequent works in the jazz genre.} \\ \cmidrule(l){2-4} 
 & Hypothesis & \multicolumn{2}{p{10cm}}{The observations are the $V_?$ such that $\exists V_1, subsequentWork(V_1, V_?) \land genre(Jazz, V_1)$} \\ \cmidrule(l){2-4} 
 & Conclusion & Extra: Here\_'Tis &  \\ \cmidrule(l){2-4} 
 & Jaccard & 0.875 &  \\ \cmidrule(l){2-4} 
 & Smatch & 0.400 &  \\ \bottomrule
\end{tabular}
\caption{DBpedia50 Case study.}
\label{tab:db-case1}
\end{table*}

\end{document}